\documentclass[11pt,a4paper]{article}
\usepackage[hyperref]{acl2021}
\usepackage{times}
\usepackage{latexsym}

\usepackage{microtype}

\usepackage{booktabs}
\usepackage{graphicx}
\usepackage{multicol}
\usepackage{multirow}
\usepackage{amsmath}
\usepackage{amssymb}

\usepackage{pifont}
\usepackage{color}
\usepackage{soul}

\aclfinalcopy % Uncomment this line for the final submission
\def\aclpaperid{1209} %  Enter the acl Paper ID here

\title{Long Text Generation by Modeling Sentence-Level and Discourse-Level Coherence}

\author{
 Jian Guan$^1$, \textbf{Xiaoxi Mao$^2$, Changjie Fan$^2$, Zitao Liu$^3$, Wenbiao Ding$^3$, and Minlie Huang}$^1$\Thanks{~Corresponding author}\\
$^1$The CoAI group, DCST; $^1$Institute for Artificial Intelligence; \\$^1$State Key Lab of Intelligent Technology and Systems;\\
{$^1$Beijing National Research Center for Information Science and Technology;}\\ 
{$^1$Tsinghua University, Beijing 100084, China.}
{$^2$Netease Fuxi AI Lab. $^3$TAL Education Group.}\\
  \small{\texttt{j-guan19@mails.tsinghua.edu.cn}, \texttt{ \{maoxiaoxi,fanchangjie\}@corp.netease.com},}\\ \small{\texttt{zitao.jerry.liu@gmail.com},} \small{\texttt{dingwenbiao@100tal.com},} \small{\texttt{aihuang@tsinghua.edu.cn}} \\
}
\date{}

\begin{document}
\maketitle
\begin{abstract}
Generating long and coherent text is an important but challenging task, particularly for open-ended language generation tasks such as story generation. Despite the success in modeling intra-sentence coherence, existing generation models (e.g., BART) still struggle to maintain a coherent event sequence throughout the generated text. We conjecture that this is because of the difficulty for the decoder to capture the high-level semantics and discourse structures in the context beyond token-level co-occurrence. In this paper, we propose a long text generation model, which can represent the prefix sentences at sentence level and discourse level in the decoding process. To this end, we propose two pretraining objectives to learn the representations %at sentence level and discourse level 
by predicting inter-sentence semantic similarity and distinguishing between normal and shuffled sentence orders. Extensive experiments show that our model can generate more coherent texts than state-of-the-art baselines.
%of the already generated prefix
\end{abstract}

\section{Introduction}

\begin{figure}[!ht]
  \centering
\includegraphics[width=\linewidth]{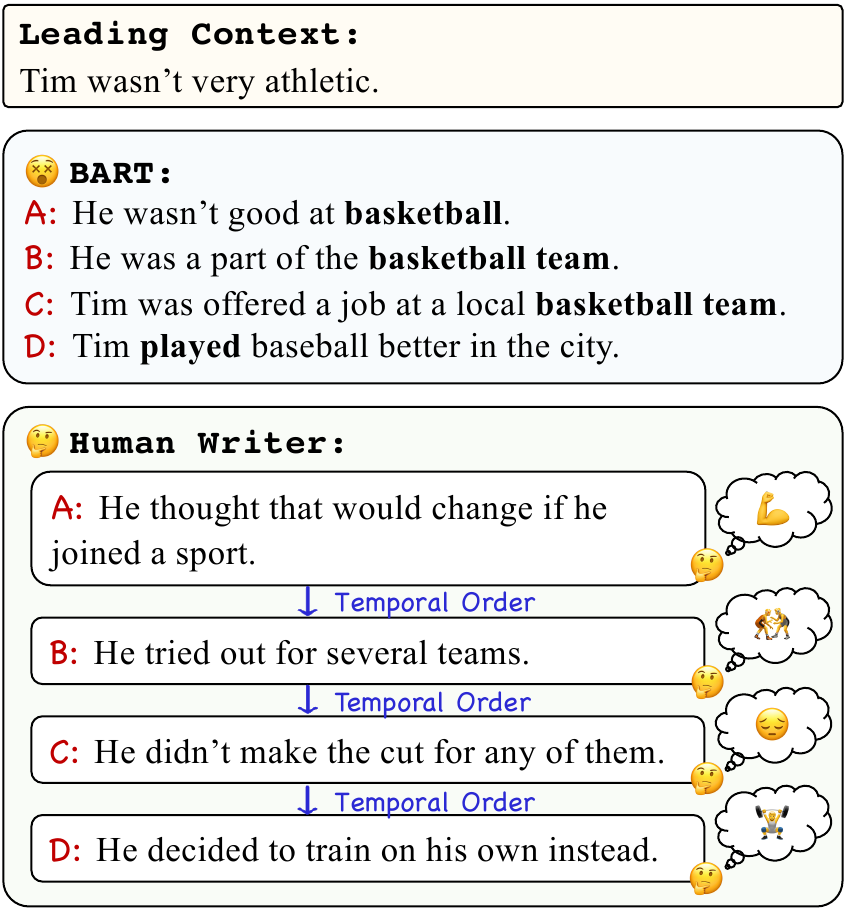}% 1\linewidth
  \caption{Story examples written by the fine-tuned BART model~\cite{bart} and a human writer given the same leading context from ROCStories~\cite{mostafazadeh2016corpus}. The generated story by BART suffers from severe incoherence issue in spite of some related concepts~(in \textbf{bold}). In comparison, the human writer can write a coherent story because they fully consider the context semantics and discourse relations~(e.g., the temporal order) among the sentences.}
  \label{fig:example}
\end{figure}

The ability to generate coherent long texts plays an important role in many natural language generation~(NLG) applications, particularly for open-ended language generation tasks such as story generation, namely generating a reasonable story from a prompt or a leading context. While existing generation models~\cite{fan2018hierarchical, radford2019language} can generate texts with good intra-sentence coherence, 
it is still difficult to plan a coherent plot throughout the text, even when using the powerful pretrained models, %such as BART~\cite{bart}, 
as illustrated in Figure~\ref{fig:example}. 

Pretrained generation models have shown state-of-the-art performance on various NLG tasks such as summarization and translation~\cite{radford2019language,bart}. However, such tasks provide sufficient source information in the input for generating desired texts, while open-ended generation tasks require expanding reasonable plots from very limited input information~\cite{guan2020knowledge}. As exemplified in Figure~\ref{fig:example}, we observe severe issues of incoherence when applying BART for story generation. Although BART performs reasonably well at generating some concepts related to the context~(e.g., \textit{``basketball'', ``player''}), they are used incoherently in the generated texts, which is manifested in repetitive plots~(e.g., the sentences \textit{B} and \textit{C}), unrelated events~(e.g., \textit{``played baseball better''}) and conflicting logic~(e.g., \textit{``not good at basketball''} but \textit{``in the basketball team''}). These issues are also commonly observed in other NLG models~\cite{holtzman2019curious,guan2020union}. We argue that existing models are rarely trained beyond the token-level co-occurrence, 
%focus overwhelmingly on learning from token-level co-occurrence 
%but lack a high-level understanding of the context, 
and therefore they can easily generate related concepts but do not arrange them reasonably. 
In contrast, human writers always first fully understand the semantics~(e.g., some key events such as \textit{``try out'', ``not make the cut''}) and the discourse relations~(e.g., temporal orders) among the already written sentences before deciding the following content. In this way, the writers can write coherent stories even with few related concepts, as shown in Figure~\ref{fig:example}. Therefore, it is important for subsequent generation to capture high-level features in the context. %the prefix information.% beyond the token-level connections across words. %And the difficulties lie in understanding the sentence-level semantic meaning and the discourse relations among already generated sentences beyond the token-level connections across words, as well as deciding the following content based on the understanding of the context.

In this paper, we propose
\textsc{Hint}, \textit{a generation model equipped with \textbf{HI}gh-level representations for lo\textbf{N}g \textbf{T}ext generation}. %, to deal with the challenge.
Typical generative models usually train a left-to-right decoder by next word prediction based on the attention to all the prefix words. In order to encourage the model to capture high-level features, we extend the decoder to represent the prefix information at sentence level and discourse level, respectively, with special tokens which are inserted at the end of each sentence.  %recent work has shown that post-training BERT~\cite{devlin2018bert} with sentence-level objectives~(e.g., natural language inference~\cite{reimers2019sentence}) or discourse-level objectives~(e.g., sentence unshuffling~\cite{lee2020slm}) has been shown helpful to improve sentence representation for understanding long texts. %We demonstrate the \textsc{Hint} approach based on BART. 
To effectively learn the representations, we propose two pretraining objectives including: %besides the general language modeling objective 
(a) \textit{semantic similarity prediction}, which requires predicting the inter-sentence similarity using the sentence-level representation, with the powerful sentence understanding model SentenceBERT~\cite{reimers2019sentence} as the teacher model; and (b) \textit{sentence order discrimination}, which requires distinguishing between the normal and shuffled sentence orders using the discourse-level representation. The objectives are designed to help the decoder capture the semantics and discourse structure of the prefix, which can benefit modeling the long-range coherence when generating long texts. We summarize our contributions in two folds:

\noindent\textbf{I.} We propose a generation model named \textsc{Hint} for long text generation. \textsc{Hint} derives high-level representations for each decoded sentence to model the long-range coherence. We adopt two pretraining objectives called similarity prediction and order discrimination to learn the representations at sentence level and discourse level, respectively.%, thereby improving the ability to understand the context and coherence of generated texts.

\noindent\textbf{II.} We conduct extensive experiments on commonsense story and fiction generation tasks. Results show that \textsc{Hint} can learn meaningful high-level representations and generate more coherent long texts than baselines.\footnote{The codes are available at \url{https://github.com/thu-coai/HINT}}
    
%always first fully understand what has been written~(e.g., the inter-sentence discourse relations) before continuing when creating a long text, as illustrated in Figure~\ref{fig:example}. In contrast, the human writers can write coherent stories 

\begin{figure*}[t]
  \centering
\includegraphics[width=\linewidth]{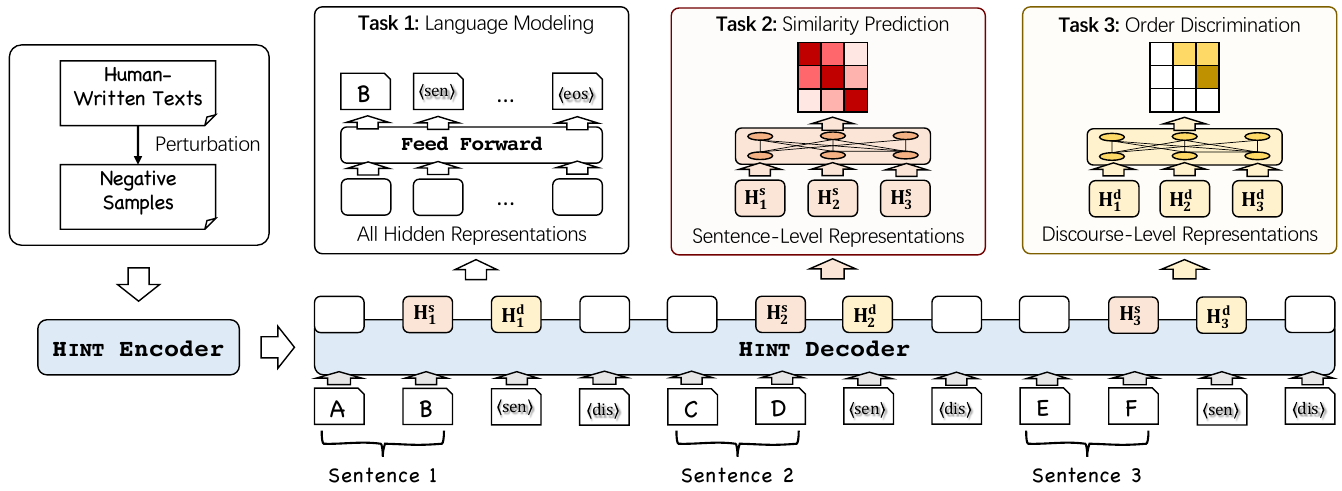}% 1\linewidth
  \caption{Model overview of \textsc{Hint}, which is pretrained to predict the next token~(Task 1), predict inter-sentence semantic similarity with the sentence-level representations~(Task 2), and distinguish between normal and shuffled sentence orders with the discourse-level representations~(Task 3) based on the human-written texts and auto-constructed negative samples.}
  \label{fig:model}
\end{figure*}

\section{Related Works}

\paragraph{Long Text Generation} Recent studies tackle the incoherence problem in long text generation from the following perspectives. \citet{li2015hierarchical} adopted a hierarchical RNN-based decoder to learn the sentence representation but without any external supervision. \citet{shao2017generating} proposed a self-attention mechanism to attend on the prefix by appending it to the RNN-based encoder, which is a similar idea with the vanilla Transformer~\cite{vaswani2017attention}. However, the token-level self-attention mechanism still struggles to model high-level dependency in the context. Recent works proposed several multi-step generation models~\cite{fan2018hierarchical,%xu2018skeleton,
yao2018plan,%fan-etal-2019-strategies,
shao-etal-2019-long,DBLP:journals/corr/abs-2006-15720,DBLP:conf/emnlp/Goldfarb-Tarrant20}, which first plan high-level sketches and then generate texts from the sketches. However, the lack of exposure to degenerate sketches may impair the generation performance since the models are only trained on sketches constructed from golden truth texts~\cite{DBLP:journals/corr/abs-2006-15720}. Another line is to incorporate external knowledge into generation especially for commonsense story generation~\cite{guan2020knowledge,DBLP:conf/emnlp/XuPSPFAC20}. However, the methods may not be always effective for other types of generation tasks.
\citet{guan2020knowledge} also required the decoder to distinguish true texts from negative samples to alleviate potential issues such as repetition. But the classification objective does not provide explicit guidance for generation at each step. %the discrepancy between the artificial simulation and actual model outputs may make the classification objective trivial and not bring substantial improvement. 
Therefore, the coherence of language generation is still an open problem.% even for the large-scale pretrained models.

\paragraph{High-Level Language Representation} Significant advances have been witnessed in many NLP tasks with pretrained contextualized representation~\cite{peters-etal-2018-deep,devlin2018bert}. However, most models were limited on token-level representation learning, which is not enough for capturing the hierarchical structure of natural language texts~\cite{DBLP:conf/acl/RibeiroWGS20}. 
%token representation is reported to perform poorly to derive semantically meaningful sentence representation~\cite{reimers2019sentence},
Several works have tried to learn high-level representation. Skip-Thought vectors~\cite{kiros2015skip} learned to encode a sentence by reconstructing its neighboring sentences. %But it consider did not consider the dynamic contextualization of sentences. 
HLSTM~\cite{yang2016hierarchical} considered a hierarchical LSTM-based encoder to learn the contextualized sentence representation by downstream classification. %or word prediction. 
HIBERT~\cite{zhang2019hibert} incorporated the hierarchical architecture to BERT~\cite{devlin2018bert} and learned sentence representation by recovering masked sentences. SentenceBERT~\cite{reimers2019sentence} derived sentence representation by fine-tuning BERT for natural language inference.  %with siamese and triplet structures. 
\textsc{Conpono}~\cite{DBLP:conf/acl/IterGLJ20} and SLM~\cite{lee2020slm} further trained BERT to understand relations among sentences at discourse level by distance prediction and sentence unshuffling, respectively. However, all these models focused on enhancing the representation of encoders for language understanding, while improving decoders by high-level representation for long text generation is yet to be well investigated.

%\paragraph{pretraining tasks} pretrain

\section{Methodology}
\subsection{Task Definition and Model Overview}
Our task can be defined as follows: given an input $X=(x_1,x_2,\cdots,x_m)$~(e.g., a beginning or a prompt), the model should generate a multi-sentence text $Y=(y_1,y_2,\cdots,y_n)$ with a coherent plot~(each $x_i$ or $y_i$ is a token). To tackle the problem, the conventional generation models such as BART commonly employ a bidirectional encoder and a left-to-right decoder to minimize the negative log-likelihood $\mathcal{L}_{LM}$ of human-written texts:%, as follows:
\begin{align}
    \mathcal{L}_{LM}&= -\sum_{t=1}^n\log {P}({y}_t|{y}_{<t}, X),\\
    {P}({y}_t|{y}_{<t}, X)&=\text{softmax}(\textbf{H}_t\boldsymbol{W}+\boldsymbol{b}),\\
    \textbf{H}_t&=\texttt{Decoder}({y}_{<t},\{\textbf{S}_i\}_{i=1}^m),\\
    \{\textbf{S}_i\}_{i=1}^m&=\texttt{Encoder}(X),
\end{align}
where %$\hat{Y}=(\hat{y}_1,\hat{y}_2,\cdots,\hat{y}_n)$ is the human-written text of length $n$ paired with the input $X$ from the training corpus, $\hat{y}_t$ is the $t$-th token in $\hat{Y}$, 
$\textbf{H}_t$ is the decoder's hidden state at the $t$-th position computed from the context~(i.e., the prefix ${y}_{<t}$ and the input $X$), and $\textbf{S}_i$ is the contextualized representation of $x_i$ acquired from the encoder, $\boldsymbol{W}$ and $\boldsymbol{b}$ are trainable parameters.

However, as aforementioned, the models often generate incoherent texts due to the decoder's inability to capture high-level features of the prefix sentences. Therefore, we extend the decoder with high-level representations to gather the prefix information. Specifically, we split the human-written texts into sequential sentences and add special tokens at the end of each sentence, which will be used to aggregate their respective semantics and their discourse relations with one another during decoding. To this end, we devise two pretraining tasks besides the standard language modeling objective, including similarity prediction and order discrimination to learn the sentence-level and discourse-level representations, respectively, as Figure~\ref{fig:model} shows. Although we only consider sentence as segments in this work, our method can be easily extended to other syntactic levels such as phrases or paragraphs.

\subsection{Sentence-Level Representation} 
Assume that the target text ${Y}$ consists of $K$ sentences, denoted from ${Y}_1$ to ${Y}_K$~(e.g., \texttt{AB} and \texttt{CD} in Figure~\ref{fig:model}). We insert a special sentence token, $\langle \texttt{sen}\rangle$, at the end of every sentence in ${Y}$, which is designed to aggregate the semantics of each sentence. Let $\textbf{H}_k^{\texttt{s}}~(1\leqslant k\leqslant K)$ denote the decoder's hidden state at the position where the $k$-th sentence token is the golden truth for next token prediction. We expect $\textbf{H}_k^{\texttt{s}}$ to be a meaningful sentence representation for $Y_k$, which means semantically similar sentences have close representations in the vector space. Since sentence representation has been well studied for language understanding with many powerful models such as SentenceBERT~\cite{reimers2019sentence}, we propose to directly transfer their semantic knowledge for our sentence representation learning. %whichb enable us to learn useful. %without training from scratch~(e.g., by auto-encoder)
Specifically, we require the \textsc{Hint} decoder to predict the similarity of any two sentences ${Y}_i$ and ${Y}_j$ only using the corresponding sentence representations $\textbf{H}_i^{\texttt{s}}$ and $\textbf{H}_j^{\texttt{s}}$, with the SentenceBERT similarity as the golden truth\footnote{The SentenceBERT similarity is computed as the cosine distance of two sentence embeddings which are derived by applying mean-pooling on the output vectors of SentenceBERT. And we normalize the results to $[0, 1]$ range by linear scaling.}. We do not directly learn the SentenceBERT representation for each sentence but the similarity score to avoid the discrepancy between different model bias. Furthermore, to alleviate the innate bias of SentenceBERT, we do not enforce \textsc{Hint} to exactly fit the golden similarity. Instead, it would be enough that the difference between the predicted score and the golden similarity is less than a margin $\Delta\in[0,1]$. Formally, the loss function $\mathcal{L}_{Sen}$ for the similarity prediction task can be derived as follows:
\begin{align}
    \mathcal{L}_{Sen}&=\frac{1}{K^2}\sum_{i=1}^K\sum_{j=1}^K\text{max}(|p_{ij}-t_{ij}|, \Delta),\label{delta}\\
    p_{ij}&=\text{sigmoid}(s_{ij}+s_{ji}),\label{sym}\\
    s_{ij}&=(\textbf{H}^{\texttt{s}}_i)^{\rm T}\boldsymbol{W}^{\texttt{s}}\textbf{H}^{\texttt{s}}_j,
\end{align}
where $t_{ij}$ is the golden similarity, $p_{ij}$ is the predicted similarity score, $s_{ij}$ is an intermediate variable to guarantee $p_{ij}$ is symmetric with respect to $i$ and $j$, $\boldsymbol{W}^\texttt{s}$ is a trainable parameter to transform the representation space of \textsc{Hint} to that of SentenceBERT. The task explicitly exerts external supervision to learn the sentence-level representation, enhancing the ability of the \textsc{Hint} decoder to fully understand the semantics of prefix sentences.

\subsection{Discourse-Level Representation}
In analogy to the sentence-level representation learning, we also insert a special discourse token, $\langle \texttt{dis}\rangle$, after every sentence and the corresponding sentence token to gather the discourse information between different sentences. Let $\textbf{H}_k^{\texttt{d}}~(1\leqslant k\leqslant K)$ denote the decoder's hidden state at the position where the $k$-th discourse token is the golden truth to be predicted. $\textbf{H}_k^{\texttt{d}}$ should be a meaningful representation which can be used to derive discourse relations with others %between the $k$-th sentence with the others
(e.g., the $k$-th sentence precedes another one in terms of the temporal order). Previous work has shown that reconstructing the correct order from shuffled sentences helps understand the discourse relations~\cite{lee2020slm}. However, the unshuffling task is not directly applicable for NLG since the decoder should learn to dynamically model the discourse structure in the decoding process rather than wait until finishing decoding the whole text. Therefore, we propose to learn the discourse-level representation in a pair-wise manner by discriminating whether the order of two sentences is correct. Formally, we minimize the cross-entropy loss $\mathcal{L}_{Dis}$ %for the order discrimination task 
as follows:
\begin{align}
    %\mathcal{L}_{Dis}&=\frac{2}{K(K-1)}\sum_{i=1}^K\sum_{j>i}^K|q_{ij}-o_{ij}|,\\
    \mathcal{L}_{Dis}&=\frac{2}{K(K-1)}\sum_{i=1}^K\sum_{j>i}^Kl_{ij},\\
    l_{ij}&=-o_{ij}\text{log}q_{ij}-(1-o_{ij})\text{log}(1-q_{ij})\\
    q_{ij}&=\text{sigmoid}\left((\textbf{H}^{\texttt{d}}_i)^{\rm T}\boldsymbol{W}^{\texttt{d}}\textbf{H}^{\texttt{d}}_j\right),
\end{align}
where $o_{ij}$ is the golden label (1 if ${Y}_i$ should precede ${Y}_j$, 0 otherwise), $q_{ij}$ is the predicted discrimination score, and $\boldsymbol{W}^{\texttt{d}}$ is a trainable parameter. Compared with the sentence-level representation $\textbf{H}_k^{\texttt{s}}$ which aggregates the semantics of a single sentence, the discourse-level representation $\textbf{H}_k^{\texttt{d}}$ focuses more on the relationship with other sentences, thereby improving \textsc{Hint}'s ability to capture the high-level features in both content and order.
%decoder to continue subsequent generation based on full .

%They contribute to generating coherent context in content and order, respectively.
%Therefore, a single  would be meaningless. %since it is necessary to know about the current order with every sentence the before making discrimination.

\subsection{Pretraining and Fine-tuning}
To learn the high-level representations more effectively, we propose to augment the training corpus by automatically constructing negative samples from the human-written texts for pretraining. Specifically, for the order discrimination task, we randomly shuffle the sentences in human-written texts as negative samples. And for the similarity prediction task, besides the negative samples with shuffled sentences, we also randomly repeat a sentence, or substitute a sentence with another from other texts as negative samples. We expect the negative samples to help enhance the generalization ability of \textsc{Hint} during fine-tuning or inference. In summary, the overall loss function $\mathcal{L}_{Pre}$ for pretraining is computed as follows:
\begin{align}
\mathcal{L}_{Pre} = \mathcal{L}_{LM}+\lambda_1\mathcal{L}_{Dis}+\lambda_2\mathcal{L}_{Sen},\label{loss}
\end{align}
where we optimize the language modeling objective $\mathcal{L}_{LM}$ only on the human-written texts,  $\mathcal{L}_{Dis}$ on the human-written texts and the negative samples with shuffled sentences, and $\mathcal{L}_{Sen}$ on all the human-written texts and the negative samples. $\lambda_1$ and $\lambda_2$ are adjustable scale factors. By pretraining with the proposed two objectives, the decoder can better capture the semantics and discourse structures in the context. And during fine-tuning, we train \textsc{Hint} only with the language modeling objective.

\section{Experiments}
\subsection{Implementation and Pretraining Dataset}
Since our approach can adapt to all the generation models with auto-regressive decoders~(e.g., GPT-2~\cite{radford2019language}, UniLM~\cite{DBLP:conf/nips/00040WWLWGZH19}, etc.), %~(e.g., GPT-2~\cite{radford2019language}, UniLM~\cite{DBLP:conf/nips/00040WWLWGZH19}, BART~\cite{bart}, etc.), 
we use BART as the base framework of \textsc{Hint}, which has been shown to have strong performance for long text generation~\cite{goldfarb2020content}. And we also provide the performance of GPT-2 widely used in the literature. Due to the limited computational resources, we follow BART$_{\rm BASE}$'s hyper-parameters %Considering the high cost of training from scratch, 
and utilize the public pretrained checkpoint to initialize \textsc{Hint}. % based on the public checkpoint of HuggingFace's Transformers~\cite{wolf2020transformers}. %\footnote{\url{https://github.com/huggingface/ transformers}}.
The batch size is set to 10 and the maximum sequence length is set to 512 for both the encoder and the decoder. %We took Adam~\cite{DBLP:journals/corr/KingmaB14} as the optimizer and set the learning rate to 3e-5. 
The margin $\Delta$ in Equation~\ref{delta} is set to 0.1 and we present the results with other settings of $\Delta$ in the appendix. Both the scale factors $\lambda_1$ and $\lambda_2$ in Equation~\ref{loss} are set to 0.1.
%Any other configuration not mentioned here is the same as the original BART model.% and is presented in the appendix.

We adopt BookCorpus~\cite{DBLP:conf/iccv/ZhuKZSUTF15} as our pretraining dataset and split each text to sentences using NLTK~\cite{DBLP:conf/acl/BirdL04}. We create the training texts by taking a sentence as the input and the following ten sentences as the target output. Besides, we construct the same number of negative samples with the human-written texts. And it is evenly possible for a negative sample to be repeated, substituted or shuffled. We pretrain \textsc{Hint} on BookCorpus for 0.1M steps. 

\subsection{Fine-tuning Setting}
We evaluate \textsc{Hint} on %widely used for open-ended generation tasks, 
ROCStories~(\textbf{ROC} for short)~\cite{mostafazadeh2016corpus} and WritingPrompts~(\textbf{WP} for short)~\cite{fan2018hierarchical}. ROC contains 98,162 five-sentence commonsense stories. We follow \citet{guan2020knowledge} to delexicalize stories in ROC by masking all the names with special placeholders to achieve better generalization. %We randomly selected 90\%/5\%/5\% stories from ROC for training/validation/test of \textsc{Hint} in the fine-tuning phase. And we regard the first sentence of each story as the input and the last four sentences as the target output.
WP originally contains %272,600/15,620/15,138 
303,358 stories paired with writing prompts, %in the training/validation/test set, 
which are usually unconstrained on writing topics. Considering that using too many examples for fine-tuning may weaken the influence of post-training,  %due to the long distance, 
we randomly selected stories from the original validation set and test set of WP for the subsequent experiments. %The average length of the prompt and story is 28.4 and 734.5, respectively, longer than the maximum sequence length of \textsc{Hint}. 
We regard the first sentence and the prompt as the input to generate a text for ROC and WP, respectively. And we only retain the first ten sentences~(split using NLTK) of the texts in WP for fine-tuning. We present more details in Table~\ref{stat}. The batch size is set to 10/4 for ROC/WP, respectively. And other hyper-parameters are the same as the pretraining phase.

\begin{table}[!ht]
%\tiny
\small
%\scriptsize
%\footnotesize
\centering
\begin{tabular}{c|cc|ccc}
%\begin{tabular}{llll}
\toprule
\textbf{Dataset}&\textbf{Input}&\textbf{Output}&\textbf{Train}&\textbf{Val}&\textbf{Test}\\
\midrule
\textbf{ROC}&14.47&56.29&88,344&4,908&4,909\\
\textbf{WP}&30.02&185.65&26,758&2,000&2,000\\
\bottomrule
\end{tabular}
\caption{The average number of tokens in the \textbf{input} and \textbf{output} in the whole dataset, and the numbers of stories for \textbf{train}ing/\textbf{val}idation/\textbf{test}.}
\label{stat}
\end{table}

\subsection{Baselines}
We compared \textsc{Hint} with the following baselines:

\noindent\textbf{Seq2Seq:} It generates a text conditioned upon the input. For better performance, 
We implement the baseline by training BART from scratch on the downstream datasets without pretraining.

\noindent\textbf{Plan\&Write:} It first plans a keyword sequence conditioned upon the input; and then generates a text based on the keywords~\cite{yao2018plan}. We implement the model based on the codes provided by the original paper.

\noindent\textbf{GPT-2} and \textbf{BART:} They are fine-tuned on the downstream datasets with the language modeling objective.

%\noindent\textbf{BART:} It is fine-tuned on the downstream datasets with the language modeling objective.

\noindent\textbf{BART-Post:} It is first post-trained on the pretraining dataset with the original pretraining objectives of BART~(text infilling and sentence permutation) for the same number of steps with \textsc{Hint}; and then fine-tuned on the downstream datasets with the language modeling objective.

\noindent\textbf{BART-MTL:} The model is trained by fine-tuning BART on the downstream datasets with multi-task learning~(MTL), including the language modeling objective and an auxiliary multi-label classification objective~\cite{guan2020knowledge}, which requires distinguishing human-written texts from auto-constructed negative samples. %that are constructed by randomly shuffling the sentences, substituting a sentence with a negatively sampled one, or repeating a sentence in an original story.

Furthermore, we conduct ablation tests by removing the proposed components respectively to investigate the influence of each component. %with the same network structure. 
Besides, we also demonstrate the adaption of our approach to general language generation models by directly fine-tuning BART and \textsc{Hint} on downstream datasets with the proposed two objectives as auxiliary tasks. %without any post-training. 
For fair comparison, we set all the pretrained models to the base version. And we also insert the sentence token and discourse token into each training text for all the baselines.

We generate texts using nucleus sampling~\cite{holtzman2019curious} with p=0.9 and a softmax temperature of 0.7~\cite{goodfellow2016deep} to balance the trade-off between diversity and fluency. And we set the probability of generating $\langle \texttt{dis}\rangle$ to 1 if the last token is $\langle \texttt{sen}\rangle$ to ensure that \textsc{Hint} can obtain the high-level representations for each sentence. And during evaluation, we remove the special tokens in the generated texts. We apply these settings to all the baselines. 
\subsection{Automatic Evaluation}
\paragraph{Evaluation Metrics} We adopt the following automatic metrics to evaluate the performance on the test sets: \textbf{(1) Perplexity~(PPL)}: Smaller perplexity scores indicate better fluency in general. We do not count the probability values at the positions where the sentence or discourse token is the golden truth.  %when computing perplexity for all the models. 
\textbf{(2) BLEU~(B-n)}: We use $n=1,2$ to evaluate $n$-gram overlap between generated texts and human-written texts~\cite{papineni2002bleu}. %We experiment with $n=1,2$.  %since BLEU scores will become extremely low with larger $n$ for open-ended generation tasks. 
\textbf{(3) Lexical Repetition (LR-n)}: The metric computes the percentage of those texts which repeat a 4-gram at least $n$ times in all the generated texts~\cite{shao-etal-2019-long}. We set $n=2$ for ROC and $n=5$ for WP.  \textbf{(4) Semantic Repetition~(SR-n)}: The metric first computes the average top-$n$ SentenceBERT similarity between any two sentences in each generated text, and then averages the results as the final score. We set $n=1$ for ROC and $n=10$ for WP. \textbf{(5) Distinct-4~(D-4)}~\cite{DBLP:conf/naacl/LiGBGD16}: We adopt distinct-4, the ratio of distinct 4-grams to all the generated 4-grams, to measure the generation diversity.
\textbf{(6) Context Relatedness:} It is a learnable automatic metric~\cite{guan2020union}. First, we train a classifier with RoBERTa$_{\rm 
BASE}$~\cite{DBLP:journals/corr/abs-1907-11692} to distinguish human-written texts and negative samples constructed by substituting words, phrases and sentences of human-written texts randomly. Then, we use the average classifier score of all the generated texts to measure the context relatedness. \textbf{(7) Sentence Orders:} In analogy to relatedness measurement, we train another classifier %based on RoBERTa$_{\rm BASE}$ 
to distinguish human-written texts and negative samples where sentences are randomly shuffled. We use the average classifier score to measure sentence orders. We train the last two metrics based on the training sets of the downstream datasets.

\begin{table*}[!ht]
%\tiny
\small
%\scriptsize
%\footnotesize
\centering
\begin{tabular}{l|c|cc|cc|c|cc}
%\begin{tabular}{llll}
    \toprule
    %\multirow{2}{*}{\textbf{Models}}&\multirow{2}{*}{\textbf{PPL}$~\downarrow$}&\multirow{2}{*}{\textbf{B-1}}&\multirow{2}{*}{\textbf{B-2}}&\textbf{Lexical}&\textbf{Semantic}&\multirow{2}{*}{\textbf{Distinct-4}}&&\multirow{2}{*}{\textbf{Relatedness}}&&\multirow{2}{*}{\textbf{Order}}\\
    %&&&&\textbf{Repetition$~\downarrow$}&\textbf{Repetition$~\downarrow$}&&&\\
    \textbf{Models}&\textbf{PPL$\downarrow$}&\textbf{B-1$\uparrow$}&\textbf{B-2$\uparrow$}&\textbf{LR-2$\downarrow$}&\textbf{SR-1$\downarrow$}&\textbf{D-4$\uparrow$}&\textbf{Relatedness$\uparrow$}&\textbf{Order$\uparrow$}\\
    \midrule
    \textbf{Seq2Seq}&18.14&0.302&0.130&0.280&0.626&0.663&0.841&0.685\\
    \textbf{Plan\&Write}&N/A&0.297&0.130&\textbf{0.201}&0.628&0.677&0.915&0.801\\
    \textbf{GPT-2}&N/A&0.305&0.131&0.331&0.636&0.684&0.919&0.813\\
    \textbf{BART}&9.83&0.307&0.133&0.307&0.635&0.699&0.916&0.816\\
    \textbf{BART-MTL}&9.68&0.312&0.137&0.271&0.629&0.683&0.945&0.820\\
    \textbf{BART-Post}&9.49&0.326&0.147&0.279&0.632&0.698&0.947&0.842\\    
    \midrule
    \textbf{\textsc{Hint}}&\textbf{9.20}&0.334&\textbf{0.154}&0.253&0.619&0.693&0.987&0.882\\
    \textbf{~~w/o Sen}&9.25&0.332&0.152&0.264&0.622&0.702&0.970&0.873\\
    \textbf{~~w/o Dis}&9.24&0.329&0.150&0.248&0.621&0.694&0.978&0.864\\
    \textbf{~~w/o Sen\&Dis}&9.45&0.324&0.146&0.277&0.634&0.686&0.937&0.847\\
    \midrule
    \textbf{{BART} w/ aux}&9.50&0.323&0.145&0.243&\textbf{0.614}&\textbf{0.710}&0.968&0.837\\
    \textbf{\textsc{Hint} {w/ aux}}&9.22& \textbf{0.335}&0.153&{0.232}&0.615&0.700&\textbf{0.989}&\textbf{0.892}\\
    \midrule
    \textbf{\textit{Golden Text}}&\textit{N/A}&\textit{N/A}&\textit{N/A}&\textit{0.058}&\textit{0.531}&\textit{0.891}&\textit{0.970}&\textit{0.903}\\
    \toprule

\end{tabular}
\caption{Automatic evaluation results on ROC. $\downarrow$ / $\uparrow$ means the lower/higher the better. The best performance is highlighted in \textbf{bold}. %And the results of golden stories are in \textit{italic}. 
\textbf{w/o Sen} and \textbf{w/o Dis} means ablating the sentence-level and discourse-level representation learning, respectively. Namely, \textbf{w/o Sen\&Dis} means post-training only with the language modeling objective. \textbf{{BART} w/ aux} and \textbf{\textsc{Hint} w/ aux} means {f}ine-tuning BART and \textsc{Hint} on the downstream dataset with the proposed objectives as \textit{aux}iliary {tasks}, respectively.}
\label{auto-eva}
\end{table*}

\paragraph{Results on ROC}
%The automatic evaluation results for ROC are shown in
We show the results on ROC in Table~\ref{auto-eva}. We do not provide the perplexity scores of Plan\&Write and GPT-2 since they do not tokenize texts %by words rather than Byte-Pair Encoding
with the same vocabulary as used in BART. 
%it is intractable to go through all the possible intermediate keyword sequences. %scores marked with N/A are not comparable with ours because the corresponding models tokenize stories by words rather than by byte pair encodings used in GPT-2.
\textsc{Hint} outperforms all the baselines in terms of perplexity, indicating the better ability to model the texts in the test set. And \textsc{Hint} can generate more word overlaps with reference texts as shown by better BLEU scores. %However, %as previous studies observed, pretraining might lead to severe lexical repetition~\cite{guan2020knowledge}.
It is accordant with the previous observation~\cite{DBLP:conf/emnlp/XuPSPFAC20} that Plan\&Write has less lexical repetition than pretraining models possibly because small models are better at learning short term statistics~(e.g., $n$-gram) but not long term dependencies. However, \textsc{Hint} improves the situation compared with GPT-2 and BART, and has less semantic repetition than all the baselines, indicating the better ability of \textsc{Hint} to capture semantic features. Besides, our approach does no harm to the generation diversity.  %\textsc{Hint} has comparable diversity with baseline models.
\textsc{Hint} also outperforms baseline models in generating related events and arranging a proper order, as shown by the higher relatedness and order scores. Furthermore, fine-tuning with the proposed objectives as auxiliary tasks can further reduce the lexical and semantic repetition, and improve the relatedness and order scores for both BART and \textsc{Hint}, suggesting the general benefit of modeling the long-range coherence at sentence level and discourse level.

Besides, the ablation test shows that the sentence-level and discourse-level representations are relatively more important to enhance the ability to generate texts with related events and reasonable orders, respectively. And both of them contribute to reducing semantic redundancy.
%\textsc{Hint} without  has higher lexical and semantic repetition, and lower relatedness score, indicating that the objective contributes to the ability to generate related plots without redundancy. When removing the order discrimination objective, the order score drops much, suggesting that the objective can enhance the ability to generate texts with reasonable order in spite of 
When post-training only with the language modeling objective, almost all the metrics drops substantially, indicating the importance to model high-level coherence. %representations are of great help to reduce redundancy, and improve the generation quality in relatedness and order. 

Furthermore, we also notice that some models achieve even higher relatedness score than the golden texts. We summarize the possible reasons as follows: (a) It is still difficult for the learned classifier to judge implicit relatedness in some golden texts, which may require a strong reasoning ability. %as illustrated in Example 1 below; 
(b) There exist some noisy texts with poor relatedness in the golden texts. %as shown in Example 2 below; 
And (c) the systems tend to generate a limited set of texts (as demonstrated by much lower distinct-4 than golden texts) with generic plots~\cite{guan2020knowledge}, which may get high relatedness scores easily. However, we believe the learnable metric is still meaningful to compare different models with similar diversity regarding the context relatedness. %We will add more analysis about the metric in our revision.

%Although Plan\&Write has less lexical repetition possibly because it is better at learning short term dependencies, .

\paragraph{Results on WP}
We present the results on WP in Table~\ref{auto-eva-wp}. We use a larger $n$ to compute the lexical/semantic repetition since we find that all the models tend to repeat similar texts easily when generating texts with hundreds of words. 
And we do not provide the relatedness and order scores because it is difficult to train satisfactory classifiers to distinguish human-written texts from negative samples well. Table~\ref{auto-eva-wp} shows that \textsc{Hint} outperforms baselines except for lexical repetition, which is accordant with the results on ROC. %the accordant conclusion with the results on ROC, 
Therefore, the high-level representations are effective for generating long texts with different lengths and domains.

\begin{table}[!ht]
\tiny
%\small
%\scriptsize
%\footnotesize
\centering
\begin{tabular}{l|c|cc|cc|c}
%\begin{tabular}{llll}
    \toprule
    \textbf{Models}&\textbf{PPL$\downarrow$}&\textbf{B-1$\uparrow$}&\textbf{B-2$\uparrow$}&\textbf{LR-5$\downarrow$}&\textbf{SR-10$\downarrow$}&\textbf{D-4$\uparrow$}\\
    \midrule
    \textbf{Seq2Seq}&129.51&0.165&0.070&0.623&0.819&0.283\\
    \textbf{Plan\&Write}&N/A&0.199&0.070&\textbf{0.524}&0.851&0.272\\
    \textbf{GPT-2}&N/A&0.200&0.073&0.655&0.883&0.287\\
    \textbf{BART}&34.42&0.205&0.075&0.620&0.854&0.291\\
    \textbf{BART-MTL}&35.71&0.198&0.076&0.654&0.846&0.305\\
    \textbf{BART-Post}&35.11&0.205&0.076&0.671&0.862&0.271\\%0.207&0.078&0.65&\\
    \midrule
    \textbf{\textsc{Hint}}&\textbf{32.73}&\textbf{0.224}&\textbf{0.084}&0.567&\textbf{0.805}&\textbf{0.313}\\
    \textbf{~~w/o Sen}&33.08&0.216&0.080&0.598&0.823&0.303\\
    \textbf{~~w/o Dis}&33.18&0.223&0.083&0.588&0.818&0.307\\
    \textbf{~~w/o Sen\&Dis}&33.71&0.207&0.076&0.610&0.845&0.280\\
    \midrule
    \textbf{\textit{Golden Text}}&\textit{N/A}&\textit{N/A}&\textit{N/A}&\textit{0.007}&\textit{0.448}&\textit{0.928}\\
    \toprule
\end{tabular}
\caption{Automatic evaluation results on WP.}
\label{auto-eva-wp}
\end{table}

\begin{table}[!ht]
%\tiny
%\small
\scriptsize
%\footnotesize
\centering
\begin{tabular}{l|lll|l}
%\begin{tabular}{llll}
    \toprule
\multirow{2}{*}{\textbf{Models}}&\multicolumn{4}{c}{\textbf{Fluency}}\\
&\textbf{Win}&\textbf{Lose}&\textbf{Tie}&\textbf{$\kappa$}\\
\midrule
\textbf{\textsc{Hint}} vs. \textbf{BART}&37.5$^{*}$&24.0&38.5&0.58\\
\textbf{\textsc{Hint}} vs. \textbf{BART-Post}&35.5$^{**}$&21.0&43.5&0.63\\
\midrule
\textbf{\textsc{Hint}} vs. \textbf{\textsc{Hint} w/o Sen}&37.0&31.0&32.0&0.68\\
\textbf{\textsc{Hint}} vs. \textbf{\textsc{Hint} w/o Dis}&33.0&25.5&41.5&0.62\\
\textbf{\textsc{Hint}} vs. \textbf{\textsc{Hint} w/o Sen\&Dis}&35.5&28.5&36.0&0.60\\
\midrule
\midrule
\multirow{2}{*}{\textbf{Models}}&\multicolumn{4}{c}{\textbf{Coherence}}\\
&\textbf{Win}&\textbf{Lose}&\textbf{Tie}&\textbf{$\kappa$}\\
\midrule
\textbf{\textsc{Hint}} vs. \textbf{BART}&54.5$^{**}$&11.0&34.5&0.59\\
\textbf{\textsc{Hint}} vs. \textbf{BART-Post}&47.5$^{**}$&21.5&31.0&0.62\\
\midrule
\textbf{\textsc{Hint}} vs. \textbf{\textsc{Hint} w/o Sen}&47.5$^{**}$&23.0&29.5&0.67\\
\textbf{\textsc{Hint}} vs. \textbf{\textsc{Hint} w/o Dis}&42.0$^{*}$&28.0&30.0&0.63\\
\textbf{\textsc{Hint}} vs. \textbf{\textsc{Hint} w/o Sen\&Dis}&55.5$^{**}$&24.0&20.5&0.58\\
\bottomrule
\end{tabular}
\caption{Manual evaluation results on ROC. The scores indicate the percentages~(\%) of \textbf{Win}, \textbf{Lose} or \textbf{Tie} when comparing \textsc{Hint} with a baseline. $\kappa$ denotes Fleiss’ kappa~\cite{Fleiss1971Measuring} to measure the inter-annotator agreement (all are \textit{moderate} or \textit{substantial}). The scores marked with $^{*}$ and $^{**}$ mean \textsc{Hint} outperforms the baseline significantly with p-value$<$0.05 and p-value$<$0.01~(sign test), respectively.}
\label{man-eva}
\end{table}

\begin{table*}[!ht]
%\tiny
\small
%\scriptsize
%\footnotesize
\centering
\begin{tabular}{l|llll|lllll}
    \toprule
\multirow{2}{*}{\textbf{Aspects}}&\multicolumn{4}{c|}{\textbf{Coherent Examples$~\downarrow$}}&\multicolumn{5}{c}{\textbf{Incoherent Examples$~\uparrow$}}\\
%&\multicolumn{2}{c|}{\textbf{Semantics}}&\multicolumn{2}{c|}{\textbf{Discourse}}&\multicolumn{4}{c|}{\textbf{Semantics}}&\multicolumn{2}{c}{\textbf{Discourse}}\\
&\textbf{Rel}&\textbf{Neg}&\textbf{Caus}&\textbf{Temp}&%\textbf{LR}&
\textbf{Rept}&\textbf{Rel}&\textbf{Neg}&\textbf{Caus}&\textbf{Temp}\\
\midrule
\textbf{Number}&563&455&476&2,376%&2,313
&3,235&3,324&3,664&394&1,795\\
\midrule
\midrule
\textbf{BART}&11.91&9.15&10.56&10.29%&8.95
&14.11&15.60&13.69&13.47&13.04\\
%\textbf{BART-Post}&11.38&8.86&10.11&9.85&14.06&15.45&13.35&13.31&12.72\\
\textbf{BART-Post}&11.46&8.86&10.21&9.94&14.06&15.45&13.35&13.31&12.72\\
\midrule
\textbf{\textsc{Hint}}&\textbf{10.90}$^{**}$&\textbf{8.50}$^{*}$&\textbf{9.68}$^{*}$&\textbf{9.50}$^{**}$%&\textbf{9.11}
&\textbf{14.74}$^{**}$&\textbf{16.32}$^{**}$&\textbf{13.96}$^{*}$&\textbf{13.68}&\textbf{13.15}\\
    \textbf{~~w/o Sen}&11.00$^{*}$&8.55$^{*}$&9.75$^{*}$&9.53$^{**}$%&8.75
    &14.04&15.43&13.29&13.59&13.04\\
    \textbf{~~w/o Dis}&10.97$^{*}$&8.52$^{*}$&9.87&9.61$^{**}$%&9.01
    &14.64$^{**}$&16.18$^{*}$&13.83$^{*}$&13.14&12.57\\
    \textbf{~~w/o Sen\&Dis}&11.41&8.84&10.16&9.89%&8.70
    &13.80&15.14&13.17&13.04&12.51\\
\bottomrule
%\multirow{3}{*}{\textbf{Models}}&\multicolumn{2}{c}{\textbf{Semantics}}&\multicolumn{2}{c}{\textbf{Discourse}}\\
%&\textbf{Weak}&\textbf{Negated}&\textbf{Temporal}&\textbf{Casual}\\
%&\textbf{Relatedness}&\textbf{Words}&\textbf{Order}&\textbf{Order}\\

%&\textbf{Weak Similarity}&\textbf{Negated Words}&\textbf{Temporal Order}&\textbf{Casual Order}\\
\end{tabular}
\caption{Perplexity scores on the coherent or incoherent examples within different aspects including %\textit{lexical repetition} (LR), 
\textit{Semantic Repetition}~(Rept), \textit{Relatedness}~(Rel), \textit{Negation}~(Neg), \textit{Causal Relationship}~(Caus) and \textit{Temporal Relationship}~(Temp). \textbf{Number} means the number of the corresponding test examples. $\downarrow$ / $\uparrow$  means the lower/higher perplexity the better. The best performance is highlighted in \textbf{bold}. $^{*}$ and $^{**}$ indicate that the corresponding model significantly outperforms BART with p-value$<$0.05 and p-value$<$0.01~(t-test), respectively.}
\label{acts_ppl}
\end{table*}

\subsection{Manual Evaluation}
%To evaluate the fluency and coherence of generated texts, 
For manual evaluation, we conduct pair-wise comparisons with two strong baseline models (BART and BART-Post), and three ablated models of \textsc{Hint}. We randomly sample 200 texts from the test set of ROC\footnote{We do not conduct manual evaluation on WP since it would be hard to obtain acceptable annotation agreement for too long texts.} and obtain 1,200 texts from the six models. For each pair of texts (one by our model and the other by a baseline, along with the input), three annotators are hired to give a preference (win, lose, or tie) in terms of fluency and coherence, respectively. We adopt majority voting to make final decisions among the three annotators. We resort to Amazon Mechanical Turk (AMT) for annotation. We follow \citet{DBLP:conf/emnlp/XuPSPFAC20} to define \textit{fluency} as a measure of intra-sentence linguistic quality and grammatical correctness, and \textit{coherence} as inter-sentence relatedness, causal and temporal dependencies. Note that the two aspects are independently evaluated. Besides, we control the annotation quality by filtering out those annotations where the annotator can not make reasonable judgments when comparing a human-written text with a negative sample. Furthermore, we also ask workers to annotate the specific errors in the generated texts. We show the annotation instruction and the error analysis of different models  in the appendix.

Table~\ref{man-eva} shows the manual evaluation results. All the results show moderate inter-annotator agreement (0.4$\leqslant\kappa\leqslant$0.6) or substantial agreement (0.6 $\leqslant\kappa\leqslant$0.8). 
%To measure the inter-annotator agreement, we calculate Fleiss' $\kappa$~\cite{Fleiss1971Measuring} for each pair-wise comparison and all the results show moderate agreement (0.4$\leqslant\kappa\leqslant$0.6) or Substantial agreement (0.6 $\leqslant\kappa\leqslant$0.8). 
%We also conducted sign test to check the significance of the differences. The results indicate that %\textsc{Hint} performs significantly better than other baselines in both aspects. More specifically, 
And we can see that \textsc{Hint} performs significantly better than baselines in coherence by capturing the high-level features, and has comparable fluency with baselines.

%post-training on knowledge bases leads to signiﬁcant improvements in grammar and logic by offering more knowledge for expanding the story plots. And multi-task learning further enhances the performance in logic and does not affect ﬂuency of generated stories.

%\subsection{Input Ranking}
%Following \citet{fan2018hierarchical} and \citet{guan2020knowledge}, we compute \textit{input ranking accuracy} to measure how strongly the generated text of a model condition on with the input. Specifically, for a golden text, we construct another 9 negative samples by replacing the input randomly each time. If the golden text has the lowest perplexity among the 10 texts by a model, it is regarded as a correct prediction. We sample 1,000 golden texts from the test set for the evaluation. Table~\ref{tab:ir} shows the results:
\iffalse
\begin{table}[!ht]
%\tiny
\small
%\scriptsize
%\footnotesize
\centering
\begin{tabular}{c|cc|ccc}
%\begin{tabular}{llll}
\toprule
\textbf{Datasets}& \textbf{ROC} & \textbf{WP}\\
\textbf{Seq2Seq}&85.3&13.9\\
\textbf{BART}&95.1&63.7\\
\textbf{BART-Post}&96.5&64.9\\
\textbf{\textsc{Hint}}&98.1&68.5\\
\end{tabular}
\caption{}
\label{tab:ir}
\end{table}

%---------Seq2Seq---BART----BART-Post---HINT---

%--ROC---85.3%-----95.1%---96.5%-------98.1%--

%---WP---13.9%-----63.7%----64.9%-------68.5%-- 

We can see that HINT outperforms baselines on both datasets. We will add the experiment and analysis in our revision.

\fi

\subsection{Language Modeling}
It is still necessary to further investigate whether the learned representations help \textsc{Hint} capture the high-level coherence better. %context semantics and discourse structures better. %For example, we expect a good language model to understand various linguistic phenomenons such as negation and temporal order. 
Therefore, we propose to evaluate the models using individual language modeling tests in different aspects~\cite{DBLP:conf/acl/RibeiroWGS20}. To this end, we construct coherent and incoherent examples based on the test set of ROC, and compute perplexity on the examples of different aspects. Specifically, we focus on the following aspects: semantic repetition, relatedness, negation, causal and temporal relationship. We select human-written texts as coherent examples and construct incoherent examples by perturbing human-written texts. For example, we select those texts with time-related words (e.g., \textit{``then''}) as coherent examples for testing in the temporal relationship. And we exchange two sequential events connected by \textit{``then''} of a human-written text or substitute \textit{``before''} with \textit{``after''} as incoherent examples of the aspect. We show more details in the appendix.% Table~\ref{acts_case}. 

We present the results in Table~\ref{acts_ppl}. \textsc{Hint} can model the context coherence better in the above aspects than baseline models~(lower perplexity on the coherent examples), and recognize the incoherent errors more effectively~(higher perplexity on the incoherent examples). By contrast, \textbf{both BART-Post and \textsc{Hint}~(w/o Sen\&Dis) achieve an overall drop of perplexity compared with BART even on the negative examples, indicating that they may still focus on capturing the token-level features}. As for the ablation study, we can see that the sentence-level representation enhances the ability of \textsc{Hint} to capture the relatedness, negation and semantic repetition, while the discourse-level representation works mainly for causal and temporal relationship. However, we also notice the insignificant improvement of \textsc{Hint} compared with BART in recognizing the unreasonable causal and temporal relationship, which may require injecting explicit inferential knowledge besides learning sentence orders. %besides learning general order.

%select stories with weak inter-sentence similarity, negated words~(e.g., ``not'', ``disappear''), time-related words~(e.g., ``then'') or causality-related words~(e.g., ``because'') from the test set of ROC as the positive examples. Accordingly, we 

\subsection{Case Study}
We present several cases in the appendix to demonstrate that \textsc{Hint} can derive meaningful sentence-level and discourse-level representations, and generate texts with better coherence than baselines with the help of the representations.

\section{Conclusion}
We present \textsc{Hint}, a generation model for long text generation, which can represent the prefix information at sentence level and discourse level in the decoding process. We propose two pretraining objectives including inter-sentence similarity prediction and sentence order discrimination to learn the sentence-level and discourse-level representations, respectively. Extensive experiments demonstrate that \textsc{Hint} can generate more coherent texts with related context and proper sentence orders than strong baselines. Further analysis shows that \textsc{Hint} has better ability of language modeling thanks to ability of modeling high-level coherence.

%As future work, we will explore to incorporate stronger hierarchical structures into model architectures and pretraining tasks for long text generation.

\section*{Acknowledgments}
This work was supported by National Key R\&D Program of China, under Grant No. 2020AAA0104500. This work was jointly supported by the NSFC projects (Key project with No. 61936010 and regular project with No. 61876096), and the Guoqiang Institute of Tsinghua University, with Grant No. 2019GQG1. We thank THUNUS NExT Joint-Lab for the support. We would also like to thank the anonymous reviewers for their invaluable suggestions and feedback.
\section{Ethics Statement}
We conduct the experiments based on two existing public datasets ROCStories and WritingPrompts, which are widely used for commonsense story generation and fiction generation tasks, respectively. Automatic and manual evaluation show that our model outperforms existing state-of-the-art models on both datasets, suggesting the generalization of our model to different domains. Besides, our approach can be easily extended to different syntactic levels~(e.g., phrase-level, paragraph-level), different model architectures~(e.g., GPT, UniLM) and different generation tasks~(e.g., dialog generation, essay generation).

We resorted to Amazon Mechanical Turk (AMT) for manual evaluation. We did not ask about personal privacy or collect personal information of annotators in the annotation process. We hired three annotators and payed each annotator \$0.05 for comparing each pair of stories. The payment is reasonable considering that it would cost average 30 seconds for an annotator to finish a comparison. %the story length and the task difficulty.

%Does the paper describe the characteristics of the dataset in enough detail for a reader to understand which populations the technology could be expected to work for?

%Do the claims in the paper match the experimental results, in terms of how far the results can be expected to generalize?

%Does the paper describe the steps taken to evaluate the quality of the dataset? For papers concerning technology that could be deployed in user-facing applications beyond the researchers:

%Does the paper describe how the technology would be deployed in actual use cases?

%Does the task carried out by the technology match how it would be deployed?

%Does the paper address possible harms when the technology is being used as intended and functioning correctly?

%Does the paper address possible harms when the technology is being used as intended but gives incorrect results?

%Does the paper address possible harms following from potential misuse of the technology?

%If the system learns from user input once deployed, does the paper describe checks and limitations to the learning process?

%Does the paper ensure that the harms identified are not likely to fall disproportionately on populations that already experience marginalization or are otherwise vulnerable?
\bibliographystyle{acl_natbib}
\bibliography{anthology,acl2021}
\appendix

\section{Implementation Details}
We implement our model based on BART$_{\rm BASE}$ and use the public checkpoint and code of HuggingFace's Transformers\footnote{\url{https://github.com/huggingface/ transformers}}. %~\cite{wolf2020transformers}. 
Both the encoder and the decoder contain 6 hidden layers with 12 attention heads. The vocabulary consists of 50,625 tokens with Byte-Pair Encoding~\cite{radford2019language}.
And we regard $\langle \texttt{mask}\rangle$ and $\langle \texttt{s}\rangle$ in the original vocabulary as the sentence token $\langle \texttt{sen}\rangle$ and the discourse token $\langle \texttt{dis}\rangle$, respectively. %~\cite{radford2019language}. 
The learning rate for both post-training and fine-tuning is 3e-5 with Adam %~\cite{DBLP:journals/corr/KingmaB14} 
as the optimizer. The Adam epsilon is 1e-6.

It cost about 32 hours for \textsc{Hint}'s post-training on BookCorpus, and 7 hours/8 hours for fine-tuning on ROC/WP, respectively. The results are based on 1 NVIDIA TITAN X GPU.

\section{Results on the Validation Set}
Besides the performance on the test set which has been reported in the main paper, we also provide the performance on the validation set of ROC in Table~\ref{auto-eva-val} for \textsc{Hint} and strong baselines.

\begin{table}[!ht]
%\tiny
%\small
\scriptsize
%\footnotesize
\centering
\begin{tabular}{l|c|c|cc|cc}
%\begin{tabular}{llll}
    \toprule
    %\textbf{Models}&\textbf{PPL$\downarrow$}&\textbf{B-1$\uparrow$}&\textbf{B-2$\uparrow$}&\textbf{LR-2$\downarrow$}&\textbf{SR-1$\downarrow$}&\textbf{Rel}&\textbf{Or}\\
    \textbf{Models}&\textbf{PPL}&\textbf{B-1}&\textbf{LR-2}&\textbf{SR-1}&\textbf{Rel}&\textbf{Ord}\\
    \midrule
    \textbf{BART}&10.04&0.315&0.301&0.634&0.924&0.821\\
    \textbf{BART-Post}&9.75&0.321&0.278&0.630&0.949&0.850\\
    \midrule
    \textbf{\textsc{Hint}}&\textbf{9.45}&\textbf{0.331}&\textbf{0.249}&\textbf{0.623}&\textbf{0.989}&\textbf{0.881}\\
    \bottomrule
\end{tabular}
\caption{Automatic Evaluation results of different models on the validation set of ROC. We do not show BLEU-2 results due to the space limitation. \textbf{Rel} and \textbf{Ord} are short for \textit{Relatedness} and \textit{Order}, respectively.}
\label{auto-eva-val}
\end{table}

\section{$\Delta$ for Sentence-Level Representation Learning}
We tune $\Delta$ in Equation 5 to investigate the influence of the margin between the predicted similarity score of \textsc{Hint} and that of SentenceBert. We present some automatic evaluation results with different $\Delta$ in Table~\ref{auto-eva-delta}. Note that we use $\Delta=0.1$ for the experiments in the main paper. We can see that a smaller $\Delta$~(e.g., 0.01) would lead to less lexical and semantic repetition but worse fluency~(indicated by higher perplexity) and context relatedness, which may be caused by the over-fitting to the model bias of the teacher model. On the other hand, a larger $\Delta$~(e.g., 0.5) would result in worse performance in almost all the metrics even than $\Delta=1.0$ (without the similarity prediction task). The result indicates that a large $\Delta$ makes the model not learn effectively from the teacher model, and impact on the representations of the model itself. By contrast, $\Delta=0.1$ would bring better overall performance.

\begin{table}[!ht]
\tiny
%\small
%\scriptsize
%\footnotesize
\centering
\begin{tabular}{l|c|cc|cc|c}
%\begin{tabular}{llll}
    \toprule
    \textbf{$\Delta$}&\textbf{PPL$\downarrow$}&\textbf{B-1$\uparrow$}&\textbf{B-2$\uparrow$}&\textbf{LR-2$\downarrow$}&\textbf{SR-1$\downarrow$}&\textbf{Relatedness$\uparrow$}\\
\midrule
0.01&10.00&0.313&0.139&\textbf{0.249}&\textbf{0.599}&0.937\\
0.05&9.78&0.316&0.140&0.264&0.610&0.962\\
0.1&\textbf{9.20}&\textbf{0.334}&\textbf{0.154}&0.253&0.619&\textbf{0.987}\\
0.2&9.67&0.326&0.146&0.273&0.628&0.975\\
0.5&9.72&0.319&0.143&0.261&0.629&0.954\\
\midrule
\textit{1.0}&\textit{9.25}&\textit{0.332}&\textit{0.152}&\textit{0.264}&\textit{0.622}&\textit{0.970}\\
    \bottomrule
\end{tabular}
\caption{Automatic Evaluation results for \textsc{Hint} with different $\Delta$. $\Delta=1.0$ means post-training ablating the sentence-level representation learning~(\textsc{Hint} w/o Sen).}
\label{auto-eva-delta}
\end{table}

\section{Manual Evaluation}

\begin{figure*}[t]
  \centering
\includegraphics[width=0.9\linewidth]{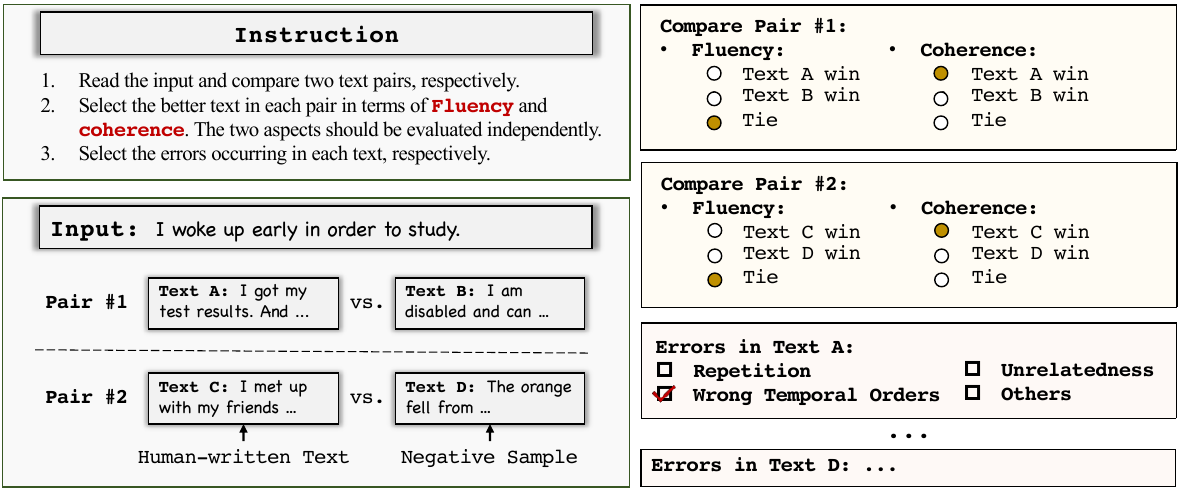}% 1\linewidth
  \caption{A simpliﬁed version of the manual annotation interface.}
  \label{fig:ant}
\end{figure*}
\paragraph{Annotation Instruction}~{}\\
We show the manual annotation interface in Figure \ref{fig:ant}. In each HIT~(human intelligence task) of AMT, we show workers an input along with two text pairs including (a) a pair of generated texts (one by \textsc{Hint} and the other by a baseline), and (b) a pair of the human-written text and a negative sample constructing by perturbing a text~(e.g., repetition, substitution) randomly sampled from the data. Note that the two pairs are presented in random order. Then, we ask workers to select the better text in each pair in terms of the fluency and coherence, respectively. Besides, we also require workers to annotate the errors in each text, including \textit{repetition}~(repeating the same or similar words), \textit{unrelatedness}~(with unrelated entities or events to the input or within its own context), \textit{wrong temporal orders}, and \textit{others}. We reject an HIT where the worker does not think the human-written text has better coherence than the negative sample, or the worker does not annotate any errors for the negative sample. In this way, we reject 21.09\% HITs in total. Finally, we ensure that there are three valid and independent comparison results for each pair of generated texts.

\paragraph{Error Analysis}~{}\\ Based on the manual annotation of errors in the generated texts, we summarize the percentages of those texts with some error in all the annotated texts~(200 for each model) in Table~\ref{tab:error}. We decide that a text contains some error when at least two of three annotators annotate the error for it. Note that each text of \textsc{Hint} is annotated five times~(three annotators each time) since \textsc{Hint} is compared with other five models. Therefore, we take the average of five annotation results. We can see that \textsc{Hint} has less repetition, better context relatedness and temporal orders than baselines. However, the results show that generating coherent long texts is still challenging.

\begin{table}[!ht]
%\tiny
\small
%\scriptsize
%\footnotesize
\centering
\begin{tabular}{lcccc}
    \toprule
\textbf{Models}&\textbf{Rept}&\textbf{Unrel}&\textbf{Temp}&\textbf{Others}\\
\midrule
\textbf{BART}&32.5&48.0&43.5&6.5\\
\textbf{BART-Post}&30.5&38.5&46.0&19.5\\
\midrule
\textbf{\textsc{Hint}}&\textbf{12.0}&\textbf{13.5}&\textbf{18.8}&\textbf{9.8}\\
\textbf{~~w/o Sen}&23.5&29.0&20.5&14.0\\
\textbf{~~w/o Dis}&16.0&15.5&42.0&18.0\\
\textbf{~~w/o Sen\&Dis}&27.5&48.5&49.0&5.0\\
\bottomrule
\end{tabular}
\caption{Percentages~(\%) of the texts which are annotated with some error in all the annotated texts.  %Proportional Distribution~(\%) of 
The error types include repetition~(Rept), unrelatedness~(Unrel), wrong temporal orders~(Temp) and others. The percentages in each row do not sum to 100\% since each text may contain multiple errors. The best performance for each error type is highlighted in \textbf{bold}.}
\label{tab:error}
\end{table}

\section{Constructing Coherent and Incoherent Examples}

Table~\ref{acts_case} presents the details for constructing examples to test the ability to model the context coherence in different aspects. However, the approach of automatic construction may inevitably introduce unexpected grammatical errors, which would also impact the text coherence. To alleviate the issue, we train a binary classifier on the CoLA corpus~\cite{warstadt2019neural} to learn to judge the grammaticality, and then filter out those examples that are classified as ungrammatical~(the classifier score less than 0.5). For simplicity, we directly use the public model from TextAttack~\cite{morris2020textattack} as the classifier, which achieves an accuracy of 82.90\% on the test set of CoLA. Finally, we filter out about 15.51\% of the test examples.%The classifier is learned by fine-tuning BERT on the CoLA corpus~\cite{warstadt2019neural}. %For simplicity, we directly use the public fine-tuned checkpoint from TextAttack~\cite{morris2020textattack}, which achieves an accuracy of 82.90\% on the test  of CoLA. %ROC would be 96.48\%, if we suppose that all of the examples are grammatical. And the accuracy is 65.68\% for WP, which is intuitive since stories in WP may contain much informal English~(e.g., website link). 

\begin{table*}[!ht]
\tiny
%\scriptsize
%\scriptsize
%\footnotesize
\centering
\begin{tabular}{p{38pt}p{150pt}p{230pt}}
\toprule
\textbf{Aspects}&{\textbf{Selecting Coherent Examples}}&{\textbf{Creating Incoherent Examples}}\\
\midrule
\textbf{Semantic\newline Repetition}&N/A&Repeating a sentence with its paraphrase by back translation\footnotemark.\newline\texttt{\textbf{Case:}} They got themselves and him on a diet. \{They put themselves on a diet with him\}$_{\rm insert}$ ...\\
\midrule
\textbf{Relatedness}&Texts with weak token-level semantic similarity in the context (e.g., with maximum inter-sentence MoverScore~\cite{zhao2019moverscore} less than 0.1). \newline\texttt{\textbf{Case:}} Lilly was afraid of heights and fast movement. She was convinced to ride a roller coaster. She hated every minute of it. She ran off and threw up immediately after ... %Lilly vowed to never go on one again. 
\textit{(Maximum inter-sentence MoverScore =0.03)}&Substituting 20\% nouns and verbs or a sentence randomly. \newline\texttt{\textbf{Case:}} The orange fell from the tree. It hit a girl on the head. \{The girl looked up at the tree.\}$_{\rm delete}$ $\rightsquigarrow$ \{She was unable to put the top up on her convertible.\}$_{\rm insert}$ Another orange fell from the tree. That orange broke her nose.\\
\midrule
\textbf{Negation}&Texts with negated words~(e.g., ``not'', ``unable'').\newline\texttt{\textbf{Case}:} The man turned it on. It \textit{did not} respond. The man unplugged it. He took it apart. He could \textit{never} get ... %that thing to work.
&Inserting or Deleting negated words for 20\% sentences.\newline\texttt{\textbf{Case}:} The man turned it on. It \{did not respond\}$_{\rm delete} \rightsquigarrow$ \{responded\}$_{\rm insert}$. The man unplugged it. He took it apart. He could never get that thing to work.\\
\midrule
\textbf{Causal\newline Relationship }&Texts with causality-related words~(e.g., ``so'', ``because'').\newline\texttt{\textbf{Case}:} Mike had a very stressful job. He needed a vacation. \textit{So} he took one. He headed to the sunny beaches of Mexico. Mike had a great time on his vacation.
&Reversing the cause and effect~(two individual sentences or clauses connected by a causality-related conjunction such as ``so''); Substituting the causality-related words with the antonyms (e.g., ``reason'' vs ``result'').\newline\texttt{\textbf{Case}:} Mike had a very stressful job. \{He took one.\}$_{\rm reverse}$ $\leftrightsquigarrow$ So \{he needed a vacation.\}$_{\rm reverse}$ He headed to the sunny beaches of Mexico ... \\%Mike had a great time on his vacation.\\
\midrule
\textbf{Temporal\newline Relationship }&Texts with time-related words~(e.g., ``then'').\newline\texttt{\textbf{Case}:} Karen got stung by a bee. Her arm swelled up immediately. It turned out she was allergic to bees! She had to go to the hospital for medication. \textit{Then} she felt much better better!&Reversing two sequential events~(two individual sentences or two clauses) connected  by a time-related conjunction; Substituting the time-related  words with the antonyms (e.g., after vs. before)\newline\texttt{\textbf{Case}:} ... Her arm swelled up immediately. It turned out she was allergic to bees! \{She felt much better better!\}$_{\rm reverse}$ $\leftrightsquigarrow$ Then \{she had to go to the hospital for medication.\}$_{\rm reverse}$\\
\bottomrule
\end{tabular} 
\caption{Instruction for selecting coherent examples from human-written texts and creating incoherent examples by perturbing human-written texts. We highlight the keywords in \textit{italic} which are crucial for the corresponding aspects. We construct the incoherent examples by inserting, deleting or reversion.}
\label{acts_case}
\end{table*}
\footnotetext[2]{The paraphrases are generated based on the public checkpoint of the back translation augmentation system of UDA~\cite{xie2020unsupervised}.}

\begin{table}[!ht]
%\tiny
%\small
\scriptsize
%\footnotesize
\centering
\begin{tabular}{p{50pt}p{53pt}p{14pt}p{20pt}p{20pt}}
    \toprule
\textbf{Sentence 1} & \textbf{Sentence 2} & \textbf{B-1}&\textbf{BART} & \textbf{\textsc{Hint}}\\
\midrule
He was really embarrassed by it. & He was very embarrassed of it. & 5.08& 2.83 & 2.17\\
\midrule
He dreamed of making the world a better place. & He had a passion to change his country for better. & 0.30&0.63 &2.17\\
\midrule
He wasn't having a good time. & He was having a good time.& 7.17&2.04&1.65\\
\midrule
I wanted to buy some fruit. & I wanted to go to a state college.&1.40&1.46&-0.50\\
\bottomrule
\end{tabular}
\caption{Sentence pairs sampled from the test set of ROC and the corresponding BLEU-1~(B-1), BART similarity and \textsc{Hint} similarity.}
\label{sen_case}
\end{table}

\begin{table*}[!ht]
%\tiny
%\scriptsize
\small
%\scriptsize
%\footnotesize
\centering
\begin{tabular}{p{50pt}p{380pt}}
\toprule
\textbf{Input:}&I woke up early in order to study.\\
\midrule
\textbf{Golden Text:}&I \textbf{met up with my friends} at the library. We \textbf{reserved a studying room} on the first floor. A few hours have passed since we \textbf{stayed for a while}. By the time \textbf{we got out}, it {was already midnight}.\\
\midrule
\midrule
\textbf{Seq2Seq:}&I am disabled and \textit{can not walk} into the library. I had to \textit{look through it for about three days}. I found out I \textit{had my book} in two days. I \textit{walked} to the library to look for it.\\
\midrule
\textbf{Plan\&Write:}&I \textit{got my test results}. And then I \textit{began to study}. I \textit{fell asleep} for the test and was very \textit{sleepy}. I {awoke to find my test results had been negative}.\\
\midrule
\textbf{BART:}&I woke up late, and \textit{found a few other people} already on the road. I was able to \textit{get to work}, but I also had to keep my eyes open. I \textit{didn't want to miss} any of the work. So I was \textit{too lazy to get to work}.\\
\midrule
\textbf{BART-CLS:}&I had a very difficult time \textit{finding my position}. I had to leave my house early in order to \textit{got to a hotel}. I had to \textit{walk around} the block in order to get to the hotel. I ended up\textit{ walking the rest} of the way to the hotel.\\
\midrule
\textbf{BART-Post:}&I drove to the library. I \textit{went to the library} and found out that the library \textit{closed} today. I \textit{got} to the library and \textit{decided to go} to the library. I \textit{sat in the library} and \textit{went to the library} and finished my studies.\\
\midrule
\textbf{\textsc{Hint}:}&I had to \textbf{go to the library} and \textbf{finish my homework}. I quickly \textbf{made it to the library} and \textbf{began to study}. After studying for an hour, I \textbf{fell asleep}. I ended up \textbf{not finishing} my homework and \textbf{went back home}.\\
\bottomrule
\end{tabular} 
\caption{Generated texts by different models. \textit{Italic} words indicate the improper entities or events in terms of coherence in the context. And \textbf{bold} words denote the coherent event sequence.}
\label{case_gen}
\end{table*}

\section{Case Study}
\paragraph{Sentence-Level Representation}~{}\\ Table~\ref{sen_case} presents some cases from the test set of ROC to demonstrate the effectiveness of the learned sentence-level representation of \textsc{Hint}. We compute BLEU-1, BART similarity and \textsc{Hint} similarity for different sentence pairs, where BART/\textsc{Hint} similarity means the cosine distance between BART/\textsc{Hint} representations of two sentences. To obtain the BART representation of a sentence, we feed it into the BART decoder~(along with its context) and apply mean-pooling on the hidden states at the last layer. \textsc{Hint} representation refers to the corresponding sentence-level representation after decoding the sentence. We normalize all the results into the standard Gaussian distribution\footnote{We compute the mean and standard deviation within 2,000 sentence pairs randomly sampled from the test set.}. We can see that \textsc{Hint} can derive meaningful sentence-level representations and gives high scores for semantically similar sentence pairs~(the first two pairs) but low scores for dissimilar pairs~(the last two pairs). By contrast, BART focuses more on token-level similarity and thus derives accordant similarity with BLEU.

\begin{table}[!ht]
%\tiny
\small
%\scriptsize
%\footnotesize
\centering
\begin{tabular}{p{184pt}}
    \toprule
%\textbf{1}&Timothy took a shower. He wiped off all the coal dust from his body. Timothy decided never to go into the mines again. He combed his hair and looked at himself in the mirror. He then saw the dirt that was still under his nails.~\texttt{\textit{(Weak)}}\\
%\midrule
%\textbf{2}&I couldn't control my anger very well. My parents would yell at me, and i ran to my room. I buried my head in a pillow and screamed. I threw my pillow and hit it hard. I tried to express my anger without them knowing.~\texttt{\textit{(Weak)}}\\
\textbf{Input: } \\\ding{172}{Kate was at her garbage can on a dark night.} \\
\midrule
\textbf{Human-written Text:}\\
\ding{173} And a raccoon was standing near the can. \\
\ding{174} It started to come towards her. \\
\ding{175} Kate turned and ran to the house hoping it wasn't behind her.\\ 
\ding{176} Once inside she was relieved to see it hadn't followed her.\\%~\texttt{\textit{(Correct)}}\\
\bottomrule
%\textbf{2}&\textbf{Kate was at her garbage can on a dark night.} Once inside she was relieved to see it hadn't followed her. It started to come towards her. Kate turned and ran to the house hoping it wasn't behind her. And a raccoon was standing near the can. ~\texttt{\textit{(Wrong)}}\\
%\midrule
%\textbf{3}&\textbf{I was riding my bike down a sidewalk.} And I heard a snapping noise. I realized I ran over a pencil. And my tube became punctured. But luckily, I was close to a bike store.~\texttt{\textit{(Correct)}}\\
%\midrule
%\textbf{4}&\textbf{I was riding my bike down a sidewalk.} But luckily, I was close to a bike store. And I heard a snapping noise. I realized I ran over a pencil. And my tube became punctured.~\texttt{\textit{(Wrong)}}\\
%\midrule
\end{tabular}

\begin{tabular}{p{30pt}p{30pt}p{25pt}p{25pt}p{25pt}}
\toprule
\textbf{Before}&\textbf{After}&\textbf{B~(M)}&\textbf{B~(D)}&\textbf{\textsc{Hint}}\\
\midrule
%\textbf{Pairs}&\textbf{1\&2}&\textbf{3\&4}&\textbf{1\&3}&\textbf{2\&4}\\ %\textbf{B-1}&\textbf{BART} & \textbf{\textsc{Hint}}\\
\textbf{}\underline{\ding{173}\ding{174}}\ding{175}\ding{176} & {\textbf{}\underline{\ding{174}\ding{173}}\ding{175}\ding{176}}&4.05&5.32&-0.89\\
\ding{173}\underline{\ding{174}\ding{175}}\ding{176} & {\ding{173}\underline{\ding{175}\ding{174}}\ding{176}}&1.30&3.81&-1.08\\
\ding{173}\ding{174}\underline{\ding{175}\ding{176}} & {\ding{173}\ding{174}\underline{\ding{176}\ding{175}}}&1.96&4.17&-3.82\\
%\textbf{1\&2}&-0.04&-0.49&1.64\\
%\textbf{2\&3}&1.58&0.92&-3.67\\
%\textbf{3\&4}&-0.01&-1.12&0.65\\
\bottomrule
\end{tabular}
\caption{A human-written text sampled from the test set of ROC with five sentences from \ding{172} to \ding{176}. We consider two adjacent sentences as a segment~(\underline{underlined}) and compute the similarity of the segment representations~(derived by BART or \textsc{Hint}) \textbf{Before} and \textbf{After} reversing the two sentences. B~(M) and B~(D) mean using BART to derive the sentence representation by mean-pooling and taking the hidden state at the position corresponding to the discourse token, respectively.}
%Based on the text, We construct three texts with wrong temporal orders by reversing two adjacent sentences~(\underline{underlined}). 
%Then, we calculate the similarity between the underlined sentences before and after reversion using the contextualized representations derived by BART or \textsc{Hint}, respectively.} %\texttt{\textit{Weak}} means that the temporal order between the sentences is not obvious in the context. Namely, the text may be still reasonable if we shuffle several sentences in it. And \texttt{\textit{Strong}} means that the text has a clear temporal order between the sentences.}
\label{dis_case}
\end{table}

\paragraph{Discourse-Level Representation}~{}\\ We also present a case in Table~\ref{dis_case} to indicate the effectiveness of the learned discourse-level representation of \textsc{Hint}. We consider a segment in the text of Table~\ref{dis_case}, which consists of two adjacent sentences~(e.g., the segment \underline{\ding{174}\ding{175}}in \ding{173}\underline{\ding{174}\ding{175}}\ding{176}). Then, we can derive the segment representation by concatenating the contextualized  representations of the two sentences. %in the normal order. 
Besides, if we reverse the two sentences~(from \underline{\ding{174}\ding{175}} to \underline{\ding{175}\ding{174}}, other sentences in the text unchanged), we can also derive the segment representation in the same way. Note that in this case we concatenate the two sentence representations still in the normal order~(i.e., first the representation of \ding{174} and then that of \ding{175}). We expect the segment representations before and after the reversion to be distant in the vector space if the sentence representation contains discourse-level information. Otherwise, the segment representations would be similar since the segments have the same tokens before and after the reversion. For BART, we derive the sentence representation by feeding the whole text into BART and mean pooling the hidden states at the positions of tokens in the sentence. And for \textsc{Hugo}, we regard the corresponding discourse-level representation of each sentence as the sentence representation. For reference, we also show the results using the hidden state of BART at the position of the discourse token as the sentence representation, i.e., B~(D). Table~\ref{dis_case} shows the similarity between the segment representations before and after the sentence reversion. All the results are normalized into the standard Gaussian distribution\footnote{We compute the mean and standard deviation within 2,000 segment pairs sampled from the test set of ROC.}.
%we can derive the contextualized representations of the two sentences by BART or \textsc{Hint}. We concatenate the two sentence representations in the normal order of them as the segment representation. Besides, we can also derive the segment representation in the same way. %We compute the similarity between texts instead of sentences since the discourse structure is reflected in the relations among different sentences. 
%We derive the contextualized representation of two sentences~(with a normal or reversed order) in a text from BART by concatenating their respective sentence representations in the normal order. For example, given the text \ding{173}\underline{{\ding{175}\ding{174}}}\ding{176}, we feed it into the BART decoder, and get the representation of \underline{{\ding{175}\ding{174}}} by concatenating the sentence representations of \ding{175} and \ding{174} in the order of {\ding{174}\ding{175}}. The sentence representation is derived by mean pooling the hidden states of all the tokens in the sentence. Analogously, we derive the \textsc{Hint} representation of two sentences by concatenating their respective discourse-level representations in the normal order. Then, we calculate the similarity between sentences before and after reversion based on the derived contextualized representations by BART or \textsc{Hint}. Note that we normalize the similarity scores into the standard Gaussian distribution\footnote{We compute the mean and standard deviation within 2,000 segment pairs. Each segment means two adjacent sentences in a text.}. 
The results show that BART derives similar representations for the segments before and after reversion whether using mean-pooling or the hidden state corresponding to the discourse token. In comparison, although the reversion does not change the sentence semantics, segment representations derived by \textsc{Hint} are very dissimilar, suggesting that \textsc{Hint} can derive meaningful discourse-level representations.

%can derive meaning discourse-level representations, 

%We obtain the BART representation of a text by applying mean-pooling on the hidden states at the last layer with the text as the input. As for the \textsc{Hint} representation of a text, we concatenate all the discourse-level representations of \textsc{Hint} for the text. We can see that BART gives accordant scores with BLEU. On the other hand, the \textsc{Hint} representations are close between two texts with discourse structures in the context~(text 1\&3, and text 3\&4) but distant between texts with different discourse structures~(text 2\&3). The result indicate \textsc{Hint} can derive meaningful discourse-level representation.

\paragraph{Text Generation}~{}\\
We presented some generated cases in Table~\ref{case_gen}. \textsc{Hint} can generate more coherent stories than baselines. Specifically, the baselines can easily predict some words which are related to the input~(e.g., \textit{``sleepy'', ``library''}) or within its own context~(e.g, \textit{``test results'', ``hotel''}). However, these words are used incoherently. For example, the text generated by Plan\&Write has a wrong temporal order among the sentences~(first \textit{``got test results''} and then \textit{``fell asleep for the test''}). The texts generated by Seq2Seq, BART and BART-CLS are chaotic in semantics and discourse structures. The text generated by BART-Post suffers from repetitive plots~(\textit{``went to the library''}) and conflicting logic~(\textit{``the library closed''} but \textit{``sat in the library''}). By contrast, the text generated by \textsc{Hint} has a coherent event sequence with related content and a proper temporal order. The results indicate the effectiveness of modeling high-level coherence for long text generation.

\end{document}